\pdfoutput=1

\documentclass[11pt]{article}

\usepackage[]{acl}

\usepackage{times}
\usepackage{latexsym}

\usepackage[T1]{fontenc}

\usepackage[utf8]{inputenc}

\usepackage{inconsolata}

\usepackage{booktabs}
\usepackage{amsfonts}
\usepackage{adjustbox}

\usepackage{subfigure}
\usepackage{amsmath}
\usepackage{graphicx}
\usepackage{hyperref}
\usepackage{arydshln}

\usepackage{pifont}

\newcommand{\rob}{RoBERTa}

\newcommand{\llama}{LLaMA}
\newcommand{\flant}{FLAN-T5}
\newcommand{\merit}{MERIt}
\newcommand{\meritp}{LogicLLM}
\newcommand{\lora}[1]{{#1}$^{\dag}$}

\newcommand{\tb}{\textbf}
\newcommand{\relto}[1]{\overset{{#1}}{\longrightarrow}}
\newcommand{\entpair}[2]{$\langle\,e_{#1},\,e_{#2}\,\rangle$}

\newcommand{\enttrips}[3]{$\langle\,e_{#1},s_{#3},e_{#2}\,\rangle$}

\title{Exploring Self-supervised Logic-enhanced Training for\\ Large Language Models}

\author{Fangkai Jiao$^{1,2}$ \qquad Zhiyang Teng$^1$  \qquad Bosheng Ding$^{1}$ \qquad Zhengyuan Liu$^2$ \\ 
\bf  Nancy F. Chen$^{1,2,}$\footnotemark[2] \qquad Shafiq Joty$^{1,3,}$\footnotemark[2] \\  
$^1$Nanyang Technological University, Singapore \\
$^2$Institute for Infocomm Research (I$^2$R), A$^*$STAR, Singapore \quad $^3$Salesforce Research \\
{\tt\small{jiaofangkai@hotmail.com \quad bosheng001@e.ntu.edu.sg \quad zhiyang.teng@ntu.edu.sg}} \\
{\tt\small{\{nfychen, liu\_zhengyuan\}@i2r.a-star.edu.sg}} \quad
{\tt\small{sjoty@salesforce.com}}}

\begin{document}
\maketitle

\renewcommand{\thefootnote}{\fnsymbol{footnote}}
\footnotetext[2]{Correspondence to: Nancy F. Chen and Shafiq Joty.}
\renewcommand{\thefootnote}{\arabic{footnote}}

\begin{abstract}
Traditional attempts to enhance the logical reasoning abilities of language models often rely on supervised fine-tuning, limiting their generalization to new tasks or domains. Large Language Models (LLMs), with their capacity to condense vast knowledge, can effectively tackle many tasks. Yet, our  experiments reveal a gap in their performance on logical reasoning benchmarks when compared to state-of-the-art fine-tuning based models. 
To bridge this gap, we present \meritp, a first-of-its-kind, fully self-supervised framework for integrating logical reasoning capabilities into LLMs, and activating them via in-context learning. We apply this to two LLM series, FLAN-T5 and LLaMA, with parameter sizes from 3 billion to 33 billion. 
\meritp~demonstrates its effectiveness through successful improvements on two logical reasoning benchmarks (ReClor and LogiQA-v2). Additionally, \meritp~based on \flant-11B attains comparable results to ChatGPT, and evaluations with \llama-based models on three  language understanding benchmarks (RACE, MMLU and Big-Bench-Hard) confirm that the improvements come without compromising the model's general language understanding capabilities.\footnote{The code and models are released at \href{https://github.com/SparkJiao/LogicLLM}{https://github.com/SparkJiao/LogicLLM}.}
\end{abstract}

\section{Introduction}
\label{sec:intro}
\begin{figure}[!t]
    \centering
    \includegraphics[width=0.95\linewidth]{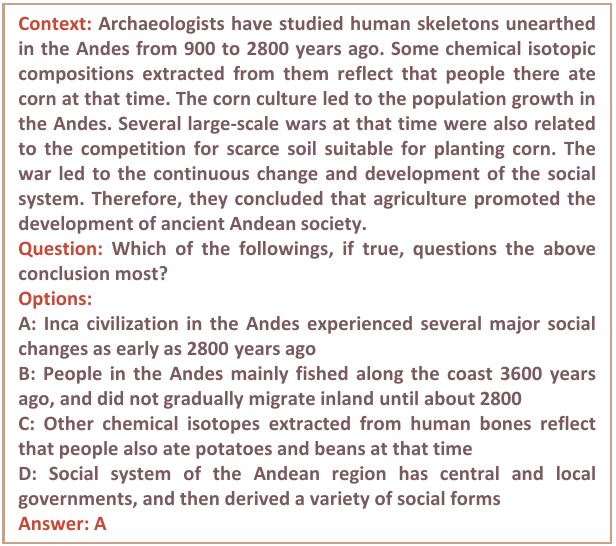}
    \caption{An example logical reasoning task from LogiQA-v2 dataset \citep{logiqa}. 
    The relations between different constituents, e.g., \textit{agriculture} and \textit{development of Andean society}, include various predicates, and it is hard to be converted into logical form through either first-order logic or formal language.
    }
    \label{fig:logiqa-sample}
    \vspace{-0.6cm}
\end{figure}
Logical reasoning serves as a bedrock for negotiation, debate and writing, underpinning our ability to engage with complex cognitive tasks~\citep{reclor}. An example of logic reasoning in natural language is shown in Figure~\ref{fig:logiqa-sample}. As the complexity of relations and expressions presented in this task defy straightforward conversion into symbolic or formal languages, perfecting logical reasoning within language models has proven to be a significant challenge~\citep{ar-lsat}.

Past attempts to incorporate logical reasoning into language models primarily focused on integrating knowledge about logic. For instance, ~\citet{dagn-huang-2021} employed graph neural networks to capture relational semantics, while ~\citet{lreasoner} used data augmentation to implement first-order logic. These techniques, however, are constrained by their need for extensive annotated training data, which hinders the model's ability to generalize across different tasks due to disparities in data distribution and optimization objectives.

Conversely, recent breakthroughs in Large Language Models (LLMs) like PaLM~\citep{palm}, LLaMA~\citep{touvron2023llama}, ChatGPT\footnote{https://openai.com/blog/chatgpt}, GPT-4 ~\cite{openai2023gpt}, and Bard\footnote{https://bard.google.com/} offer a promising alternative. These LLMs effectively encapsulate a vast array of knowledge and tackle diverse tasks with minimal specialization, guided by human instruction. Despite their potential, our  experiments on logical reasoning benchmarks revealed deficiencies in their logical reasoning capabilities as shown later in our experiments.

Contemporary efforts to fortify LLMs' specific capabilities fall broadly into two categories. The first employs external tools or APIs~\citep{toolformer,arm-survey-lecun,binder2022cheng,pal2022gao,pot2022chen}, aiding LLMs in argument parsing and semantic understanding. Yet, these tools' utility for logical reasoning remains limited due to the absence of a symbolic language for problem descriptions. The second category, instruction tuning, relies on data augmentation or enriched human feedback but struggles due to the scarcity of task-specific data and high annotation costs~\citep{instructgpt,wizardlm2023xu}.
In this work, we pivot away from these traditional methods and introduce \meritp, which performs self-supervised logic-enhanced meta-training for LLMs. It tackles two primary challenges: 1) synthesising logic-consistent data from raw texts ensuring fully self-supervised training, and 2) effectively incorporating logic prior into LLMs while preventing learning problems, such as memorization, forgetting and generalization.

To tackle the first challenge, \meritp~emphasizes the necessity of understanding and exploiting \textbf{fuzzy logical consistency}. As mentioned previously, strict formal logic is often absent in natural language, we instead treat the \textbf{relational consistency} between different perspectives of relational expressions as an approximation to fuzzy logic consistency\footnote{In this paper, we will use the term \textbf{logical consistency} to represent \textbf{consistency in fuzzy logic} for simplification, which is further described by relational consistency. This means that the relationship between a \textbf{logically consistent} data pair has a higher degree of logical consistency but cannot be strictly proved considering the diverse expressions of relations.}.
In fact, ensuring logical consistency in a discourse is a key requirement for text coherence and effective information conveyance \cite{speech-and-lang-process}. 
We devise a method that inspects the implicit intra-sentence relation of entity pairs at the discourse level to extract logically consistent examples from Wikipedia articles (Figure~\ref{fig:merit}). Specifically, we posit that direct and indirect relations of an anchor entity pair should be logically consistent, as they are derived from the ``same'' context.
For the second challenge, \meritp\ adopts an auto-regressive objective optimizing on the logically consistent relation instances directly to make it seamlessly adapt to its pretraining objective. It tasks the model with generating the alternative perspective (indirect or direct) given a direct or indirect description of the anchor entity pair. We further employ counterfactual data augmentation through entity replacement to enforce relation-centric reasoning, which not only avoids the model's tendency to merely recall results from memory but also ensures the preservation of the logic-enhanced aspect of the learning process.

\meritp\ is task-agnostic and does not require any annotations, making it adaptable to various logical reasoning tasks. We have conducted experiments across two distinct LLM series, FLAN-T5~\citep{flan-v2} and LLaMA~\citep{touvron2023llama}, encompassing a variety of parameter sizes. These experiments are designed to investigate two main questions: (1) Can the logical reasoning capabilities be exclusively improved through self-supervised meta-training for LLMs, thereby circumventing the need for task-specific supervised fine-tuning? (2) How does the logic-enhanced meta training affect the LLM's  language understanding capabilities, i.e., does it suffer from forgetting or generalization issues?


In response to the first question, our findings suggest that LLMs trained with the \meritp~objective demonstrate superior performance on logical reasoning benchmarks, eliminating the need for further fine-tuning. Our \meritp~based on \flant-11B attain comparable results to ChatGPT on two logic reasoning benchmarks, ReClor~\citep{reclor} and LogiQA-v2~\citep{logiqa2}, highlighting the feasibility of enhancing logical reasoning abilities through self-supervised training alone.

Regarding the second question, our evaluations with \llama-based models on three general language understanding benchmarks - RACE~\cite{race}, MMLU~\citep{mmlu2021} and BIG-Bench-Hard (BBH)~\citep{big-bench-hard}, confirm that the enhanced logical reasoning capabilities do not compromise the model's overall language understanding on MMLU and BBH. In fact, the learned logic ability appears to boost the model's performance in RACE.

\section{Related Work}
\label{sec:related_work}


\subsection{Large Language Models}

In recent years, Large Language Models with in-context learning have emerged as a groundbreaking paradigm in the field of NLP. Unlike the traditional fine-tuning approach, in-context learning leverages natural language instructions or a small number of annotated examples as demonstrations to predict responses for new instances. This unique approach empowers LLMs to serve as a versatile tool for handling multiple tasks without requiring task-specific training.
However, recent evaluations of LLMs~\citep{qin2023chatgpt-eval,bang2023chatgpt-eval,mt-chatgpt2023,maruf2023eval,seaeval2023} have revealed a limitation in their ability to learn complex skills like logic and planning through language modeling alone. To address this, even the training of GPT-4 has incorporated labeled matching datasets to enhance its performance in solving math word problems~\citep{openai2023gpt}.
Nevertheless, due to the vast amount of data used in pre-training LLMs, annotated data for specific capabilities may be severely undersampled, and the cost of obtaining annotations should not be overlooked. Therefore, it remains crucial to develop various self-supervised or weakly-supervised training methods that do not rely on human annotation. These approaches are essential for constructing more robust and versatile LLMs that can perform a wider range of tasks with higher proficiency and lower resource.

\subsection{Reasoning in Natural Language}


Previous research aimed at natural language reasoning tasks can be broadly classified into three categories. The first category involves explicit prior knowledge, such as discourse structure or linguistic knowledge, to model implicit reasoning processes~\citep{discern-gao-2020,dagn-huang-2021}. 
The second category is neural-symbolic reasoning, where variables are first parsed, and then predefined programs are executed to obtain final results~\citep{lreasoner,ar-lsat}. However, a significant challenge with these methods is the requirement of a robust semantic parser and a self-contained symbolic system for extracting variables or arguments, which is impractical for logic reasoning based on natural language.
The third category encompasses methods that focus on general domain pre-training for reasoning via denoising auto-encoding~\citep{rept,reason-bert,kmlm}.
Nevertheless, restricted by the poor task generalization of discriminative models with few parameters, these methods are still in demand of task-specific fine-tuning to activate learned knowledge.

Our approach in this paper falls within the third category, which improves the efforts of MERIt~\citep{merit} by transforming it into auto-regressive framework to better align the nature of LLMs as generative model. We also drop the usage of knowledge graph enabling enhancing the logic of LLMs through purely self-supervised learning.

\begin{figure*}
    \centering
    \includegraphics[width=0.85\textwidth]{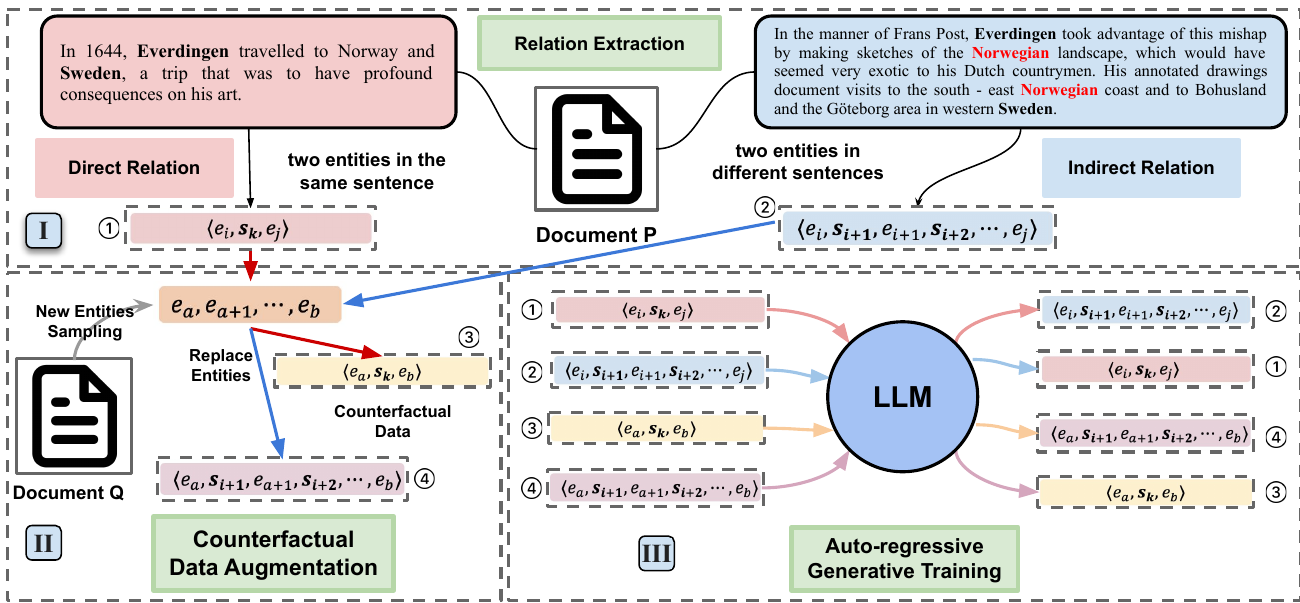}
    \caption{The \meritp~framework. $P$ and $Q$ are two arbitrary paragraphs  from Wikipedia. In Step 1, we extract intra-sentence relations \ding{172}:~\enttrips{i}{j}{k}, and the compositions of them \ding{173}:~$\langle e_i, s_{i+1}, e_{i+1}, \cdots,s_{j},e_j\rangle$ from $P$ for an entity pair \entpair{i}{j}; \ding{172} and \ding{173} are direct and indirection relations, respectively. Here $s_k$ is a relation, represented by the sentence that mentions \entpair{i}{j}.
    \ding{172} and \ding{173} are viewed as logically consistent since both of them describe the ``same'' relation between \entpair{i}{j} from different view.
    In Part I of the figure, $e_i$ refers to \textit{Everdigen} and $e_j$ represents \textit{Sweden}. The intermediate entity is \textit{Norwegian} here. The direct relation on the left says that \textit{Everdigen} has traveled to \textit{Sweden}, and the indirect relation implies the fact that \textit{Everdigen} has probably visited \textit{Sweden} as well as its nearby area, otherwise he could not complete the sketches of \textit{Norwegian}, demonstrating the fuzzy logic consistency with high probability.
    Step 2  is the process of counterfactual data augmentation, where counterfactual relation composition is generated by random entity replacement. \ding{174} and \ding{175} are the counterfactual augmentations of \ding{172} and \ding{173}, respectively. 
    Finally, in Step 3, the LLM is optimized to generate direct/indirect relations with their logically consistent indirect/direct counterparts as inputs. Here, \ding{172}$\rightarrow$ \ding{173}, \ding{173}$\rightarrow$ \ding{172}, \ding{174}$\rightarrow$ \ding{175}, and \ding{175}$\rightarrow$ \ding{174} are considered. 
    }
    \label{fig:merit}
\end{figure*}
\section{LogicLLM}
\label{sec:method}
Figure~\ref{fig:merit} shows the framework of LogicLLM. It involves three main steps: 1) Logic-consistent Data Construction (Section \ref{sec:logicdata}), which synthesises the logic-consistent data using relation discrimination between entity pairs; 2) Counterfactual Data Augmentation (Section ~\ref{sec:cfaug}), which augments the logic-consistent training data by entity sampling and replacement; 3) LLM Training (Section ~\ref{sec:llmtrain}), which performs continual training of LLMs using the training data generated by the previous two steps. 

\subsection{Logically consistent Data Construction}
\label{sec:logicdata}

Ensuring logical consistency in discourse and pragmatics is a fundamental prerequisite for natural language to effectively convey information and maintain coherence. Consequently, logically consistent data is prevalent in text documents and various techniques can be applied to extract them. 
In this study, we implement this by inspecting intra-sentence relation of entity pairs at the discourse level to extract logically consistent examples from Wikipedia.

\noindent\textbf{Direct relation} Given an arbitrary paragraph and an anchor entity pair \entpair{i}{j}, \textbf{we assume there exists an \emph{implicit} relation $s_k$ between \entpair{i}{j} if one sentence directly mentioning them can be found.} This comes from the distant supervision~\cite{mintz2009distant} and has been employed and extended in self-supervised training by previous work~\citep{reason-bert}. For example, the instance \ding{172} in Figure~\ref{fig:merit} is a direct relation. To this end, we simply treat \enttrips{i}{j}{k} as the direct relation triplet for further data construction.

\noindent\textbf{Indirect relation}  Entities $e_i$ and $e_j$ can be indirectly connected through multiple sentences within the input paragraph. In such situations, we identify a chain of triplets, such as $\langle e_i, s_{i+1}, e_{i+1}, \cdots,s_{j},e_j\rangle$, which represents an indirect relation between the entity pair \entpair{i}{j} through the relation composition of serial relation triplets \enttrips{i}{i+1}{i+1}, \enttrips{i+1}{i+2}{i+2}, $\cdots$, \enttrips{j-1}{j}{j}. For example, instance \ding{173} in Figure~\ref{fig:merit} demonstrates an indirect relation.\footnote{In practice, we find 87\% indirect relations are composed of two relation triplets, 12\% contain three triplets, and less than 1\% have more than 4 triplets. This prevents the logical consistency be weakened by long context.}

\noindent\textbf{Logical consistency} Intuitively, the direct and indirect relations between \entpair{i}{j} should be logically consistent since they are derived from same context and describing the same entity pairs. 
Instances \ding{172} and \ding{173} in Figure~\ref{fig:merit} exemplify logically consistent relations. By establishing implicit connections between single-step and multi-hop reasoning, LLMs gain the ability to understand relation composition process between $s_k$ and $\langle s_{i+1}, s_{i+2}, \cdots, s_{j-1}\rangle$. This capability consequently enhances the LLMs' logical reasoning abilities.

To retrieve logically consistent relation pairs, we follow a two-step process. First, we recognize all entities  within each paragraph via distant annotation from WikiData~\citep{wikidata}. And secondly, we enumerate every possible entity pair and search for a series of sentences and check if both direct and indirect relations can be extracted.

\subsection{Counterfactual Data Augmentation}
\label{sec:cfaug}
The work we have described in Section \ref{sec:logicdata} produces logically consistent data that correlates entities and relations within reasoning paths. To enhance entity-irrelevant reasoning and ensure LLM focuses more on the process of relational composition rather than the entities themselves, we have additionally introduced counterfactual data augmentation. This approach, similar to the method suggested by \citet{merit}, includes the random replacement of entities.

To create counterfactual examples of \entpair{i}{j} within paragraph $P$, we initially select a random paragraph, denoted as $Q$, from a separate document. Subsequently, we sample a new set of entities, such as $e_a, e_{a+1}, \cdots, e_b$ from $Q$. The head and tail entities in the original relation instances of \entpair{i}{j} are then substituted by these randomly sampled entities, maintaining the relationships unchanged. For instance, after substituting $e_i$ and $e_j$ with $e_a$ and $e_b$, \ding{174} and \ding{175} become the counterfactual augmentations of \ding{172} and \ding{173}, respectively. In our research, we postulate that the logic-consistency between $s_k$ and $s_{i+1}, e_{i+1}, s_{i+2}, \cdots, s_{j-1}$ remains undisturbed in the counterfactual examples. This assertion is based on the idea that logical relationships within a paragraph's context are primarily driven by shared entities and their interconnections rather than the specific entities themselves. 

\subsection{Training Objective}
\label{sec:llmtrain}
During the training phase, we apply continual training to LLMs  using logic-consistent data. 
Drawing inspiration from the success of in-context learning, 
we treat one relation from a logic-consistent relation pair as the in-context example and task the LLM with generating the other relation. As depicted in Figure~\ref{fig:merit}, using the logic-consistent pair $\langle$\ding{172}, \ding{173}$\rangle$ as an example, when \ding{172} is given as the conditional input, the LLM is expected to produce \ding{173} as the output, and vice versa. This process intuitively forces the LLM to reason the logic-consistent connections between the input and output relations since they are from the same context and the entity pairs of \ding{172} and \ding{173} are both $e_i$ and $e_j$.

Formally, we denote the data extracted from Section~\ref{sec:logicdata} and Section~\ref{sec:cfaug} as $D = \{\langle R_{i}^{1}, R_{i}^{2} \rangle \}_{i=1}^{N}$, where $N$ represents the number of training examples, and $\langle R_{i}^{1}, R_{i}^{2} \rangle$ is the $i$-th logic-consistent record. Here, $R_{i}^{1}$ refers to the direct relation-related instance, while $R_{i}^{2}$ represents the instance with an indirect relation. The goal of LLM training is to minimize the negative log-likelihood function as follows:
\begin{equation}
\label{eqn:logic}
\small{
\begin{aligned}
    \mathcal{L}_{\mathrm{logic}}&=-\sum_{i=1}^{N} [\log P(R_i^1|R_i^2)  + \log P(R_i^2|R_i^1)]\\
                             &=-\sum_{i=1}^{N} [\sum_{j=1}^{|R_i^1|}\log P(R_{i,j}^1 |R_{i,<j}^{1},R_{i}^2) \\
                             &\ \ \ \ \ \ \ \ \ \ \ \ \ \ \ \ + \sum_{j=1}^{|R_i^2|}\log P(R_{i,j}^2 |R_{i,<j}^{2},R_{i}^1)],
\end{aligned}
}
\vspace{-0.3cm}
\end{equation}
where $R_{i,j}^1$, $R_{i,j}^2$ denotes the $j$-th token of $R_{i}^1$ and $R_{i}^2$, respectively.

Furthermore, we incorporate the another causal language modeling loss $\mathcal{L}_{\mathrm{lm}}$ to mitigate the catastrophic forgetting problem. Both $\mathcal{L}_{\mathrm{lm}}$ and $\mathcal{L}_{\mathrm{logic}}$ are implemented as auto-regressive decoding. The only difference is that they sample from different data source. $\mathcal{L}_{\mathrm{lm}}$ continuously samples data from the subset of training corpus used during the last-stage pre-training, i.e., Wikipedia paragraphs for \llama~series models, and FLAN-collection-v2 for \flant~series models.
Therefore, the overall training objective is defined as: 
\begin{equation}
\label{eqn:overall}
\small
\mathcal{L}=\mathcal{L}_{\mathrm{logic}}+\mathcal{L}_{\mathrm{lm}}.
\end{equation}
During training, for each forward-backward, we randomly sample two mini-batches with the same size from the datasets for logic-enhanced training and language modeling, respectively, and merge them into a single one.

\section{Experiment}


We integrate our pre-training approach into two prominent LLMs: LLaMA~\citep{touvron2023llama} and FLAN-T5~\citep{flan}. These models boast parameter sizes ranging from 3 billion to 30 billion.
To thoroughly evaluate the capability of LLMs from various angles, we have carefully selected five datasets representing three distinct categories. 
ReClor~\citep{reclor} and LogiQA-V2~\citep{logiqa} are two logical reasoning benchmarks sourced respectively from standardized graduate admission examinations and logical examination papers intended for reading comprehension.
RACE~\citep{race} is a reading comprehension task that assesses general reasoning abilities. 
MMLU~\citep{mmlu2021} is used for measuring the learned knowledge and massive multitask language understanding, and BIG-Bench-Hard (BBH)~\citep{big-bench-hard} is a collection of multiple challenging tasks where LLMs fall behind human being.
By employing MMLU and BBH, we aim to verify whether the logic-oriented meta-training negatively impacts the models' ability to generalize across a wide range of tasks. Due to space limitation, more implementation details can be found in Appendix~\ref{sec:app-imple}.

\section{Results and Analysis}

\begin{table}[!t]
\centering
\setlength{\tabcolsep}{2.5mm}{
\scalebox{0.8}{
\begin{tabular}{lcccc}
\toprule
                        & \multicolumn{2}{c}{\textbf{ReClor}}  & \multicolumn{2}{c}{\textbf{LogiQA-v2}} \\
Model / Dataset         & Dev         & Test             & Dev     & Test           \\
                        & Acc.        & Acc.             & Acc.    & Acc.           \\ \hline
ChatGPT                 & 56.6        & 61.2        & 54.5              & 52.7                 \\ 
\hline
\llama-7B               & 30.2        & 30.3        & 27.4              & 28.1            \\
~~~~w/ \meritp           & \bf{32.4}   & \bf{31.0}   & \bf{27.7}         & \bf{28.6}     \\ 
\hdashline
\llama-13B              & 30.4        & 33.5        & 33.0             & 32.1               \\
~~~~w/ \meritp           & \bf{37.4}   & \bf{36.3}   & \bf{34.1}        & \bf{34.0}    \\ 
\hdashline
\llama-33B              & 45.2          & 50.3       & 41.2              & 41.6          \\
~~~~w/ \lora{\meritp}   & \bf{50.2}     & \bf{54.4}  & \bf{45.9}         & \bf{42.6} \\
\hline 
Falcon-40B              & 38.4          & 37.1       & 35.9             & 36.1  \\
~~~~w/ \lora{\meritp}   & \bf{41.4}     & \bf{43.0}  & \bf{38.6}        & \bf{37.2}      \\\hline
FLAN-T5-3B              & 54.6        & 52.5        & 48.7              & 48.7      \\
~~~~w/ \meritp~\& FLAN     & \bf{55.8}   & \bf{54.1}        & \bf{50.8}   & \bf{50.1} \\ 
  \hdashline
FLAN-T5-11B             & 57.4        & 59.9             & 55.3        & 53.1   \\  
~~~~w/ \meritp~\& FLAN     & \bf{61.2}   & \bf{61.1}        & \bf 56.0        & \bf{54.0}\\  
\bottomrule
\end{tabular}
}}
\caption{The results on logical reasoning benchmarks. Better results are annotated in bold. \lora{} refers that the corresponding model is trained through QLoRA~\citep{dettmers2023qlora}.
}
\label{tab:logic}
\end{table}

\subsection{Logical Reasoning}
\label{sec:logic-reason}
Table~\ref{tab:logic} shows the results on ReClor and LogiQA-v2 under zero-shot setting.
%
%
From the table we can find that the performance of \llama-based models is notably lower compared to ChatGPT. 
By training \llama~models with \meritp, we observe significant enhancement in their zero-shot logical reasoning capabilities. For instance, on \llama-13B and \llama-33B, the average improvements across the four dataset splits are 3.2 and 3.7 points, respectively. The benefits are more substantial than those observed in the 7B models (0.9 points), which aligns with the findings on emergent abilities \citep{emergent2022jason}. This could be attributed to the fact that larger models possess stronger generalization abilities and better apply their learned capabilities to different tasks. 
We also conducted experiments on Falcon-40B~\citep{Falcon}, and found that \meritp~brings an average improvement of 3.2 points.


%
Consistent with \llama-based models, we can draw similar conclusions for those based on \flant, where logic-oriented meta-training also yields improvements for both \flant-3B and \flant-11B.
For \flant-11B, our model achieves accuracies of 61.2 and 61.1 on the development and test sets of ReClor, respectively. On the development and test sets of LogiQA-v2, our logic-oriented \flant-11B model achieves accuracies of 56.0 and 54.0, respectively. Notably, on the development set of ReClor, our logic-oriented \flant-11B model outperforms ChatGPT by a significant margin of 4.8 accuracy points. Similarly, on the development and test sets of LogiQA-v2, our logic-oriented \flant-11B model surpasses ChatGPT by 1.5 and 1.3 accuracy points, respectively.
These overall results indicate that instruction tuning on multiple supervised datasets, such as the FLAN collection, can still be improved for learning logic. We hypothesize that this may be attributed to the sparsity of reasoning-relevant data in the entire collection and the conflicts between different tasks. 

\begin{table}[t]
\centering
\setlength{\tabcolsep}{2.5mm}{
\scalebox{0.85}{
\begin{tabular}{lccccccc}
\toprule
                              & \multicolumn{2}{c}{\textbf{RACE}} & \multicolumn{2}{c}{\textbf{MMLU}} \\
Model / Dataset                     & Dev       & Test      & 0-shot      &  5-shot \\
                                    & Acc.      & Acc.      & Acc.        &  Acc. \\ \hline
\llama-7B                     & 31.3      & 32.3            & 33.3          & 36.2\\
~~~~w/ \meritp                & \bf{37.3} & \bf{37.9}       & \bf{34.6}     & \bf{36.6} \\ 
\hdashline
\llama-13B                    & 55.8      & 54.5         & 41.1       & 46.7 \\
~~~~w/ \meritp                & \bf{57.7} & \bf{55.6}   & \bf{43.3}  & \bf{47.3} \\ 
\hdashline
\llama-33B                   & 68.4      & \bf{68.1}     & 54.3      & \bf 58.3 \\
~~~~w/ \lora{\meritp}        & \bf{68.8} & \bf{68.1}    & \bf{54.4} & \bf 58.3 \\ 
\bottomrule
\end{tabular}
}}
\caption{The results of \llama~models on RACE and MMLU. $^\dag$ means training through QLoRA. 
}
\label{tab:dream}
\end{table}

\subsection{Hybrid Reasoning and Application}

In addition to logical reasoning in text, we are also curious about whether logic-enhanced training contributes to general language understanding (RACE), and maintain the general capabilities on massive knowledge based tasks (MMLU). To investigate this, we evaluate the performance of the enhanced \llama~models on these two datasets.
 
As shown in Table~\ref{tab:dream}, 
from 7B to 33B, \meritp~can consistently improve the performance on RACE, except the one of \llama-33B w/ \meritp on the test set.
Specifically, \llama-7B w/ \meritp~obtain around 4.2 absolute improvements, and \llama-13B w/ \meritp~achieves 1.5 improvements, which has verified that the logic-enhanced training is also beneficial to general reasoning and reading comprehension.
Additionally, we find that \meritp~can also benefits the massive multitask language understanding (MMLU) on \llama-7B and 13B.
We find that the improvements of both RACE and MMLU on \llama-33B are marginal, probably because low-rank adaptation have restricted the generalization.


\begin{table}[t]
\centering
\setlength{\tabcolsep}{2.5mm}{
\scalebox{0.8}{
\begin{tabular}{lcccc}
\toprule
                        & \multicolumn{2}{c}{\textbf{ReClor}}  & \multicolumn{2}{c}{\textbf{LogiQA-v2}} \\
Model / Dataset                 & Dev         & Test             & Dev     & Test           \\
                                & Acc.        & Acc.             & Acc.    & Acc.           \\ \hline
\llama-13B                      & 30.4        & 33.5        & 33.0             & 32.1               \\
~~~~w/ \meritp~(\textit{ctr})    & 33.4        & 33.3        & 33.1             & 32.7    \\ 
~~~~w/ \meritp~(\textit{ar})     & \bf{37.4}   & \bf{36.3}   & \bf{34.1}        & \bf{34.0}    \\ 
\hdashline
\llama-33B                      & 45.2          & 50.3       & 41.2              & 41.6               \\
~~~~w/ \lora{\meritp}~(\textit{no aug.}) & 49.4          & 53.0       & 44.2              & 40.8    \\ 
~~~~w/ \lora{\meritp}~(\textit{1 aug.})	 & 50.8	         & 52.7	      & 45.6	       &  41.5     \\
~~~~w/ \lora{\meritp}           & \bf{50.2}     & \bf{54.4}  & \bf{45.9}         & \bf{42.6}    \\ 
\bottomrule
\end{tabular}
}}
\caption{The effect of different training objectives. \textit{Ctr} refers contrastive learning and \textit{ar} means the auto-regressive variant. \textit{no aug.} means the counterfactual data augmentation is removed from the \meritp~framework. \lora{}~means that the model is trained with QLoRA.}
\label{tab:ablation}
\end{table}



\subsection{Pre-training Strategy}

\meritp\ draws inspiration from the contrastive learning framework for logical reasoning, i.e., \merit, which has demonstrated its efficacy in fine-tuning based approaches. As mentioned earlier, we hypothesize that contrastive learning may be inadequate for LLM with in-context learning. To validate this assumption, we examine the effects of contrastive learning (\textit{ctr}) and auto-regressive generation (\textit{ar}). In the case of contrastive learning, we adopt the methodology of \merit\ to construct logically inconsistent instances and optimize the model by maximizing the distance between logically consistent instances and the inconsistent counterparts.
Referring to the table, it can be observed that \meritp~(\textit{ctr}) fails to yield significant improvements compared to \llama-13B, except for the dev set of ReClor. Conversely, the auto-regressive models consistently outperform both the baseline models and the contrastive methods by considerable margins across all dataset splits.
We propose two primary reasons to explain the superiority of auto-regressive models over the contrastive approach.

First, the heuristic construction process for negative candidates used in contrastive learning fails to identify true contradictory relations, resulting in randomly chosen negative samples that lack logically opposite relationships with the positive instances. 
To this end, the contrastive learning process can degrade into a positive-only optimization process, which is similar to auto-regressive learning but receives less token-level supervision.

Second, the divergence between the training objectives of contrastive learning and auto-regressive generation undermines the model's ability to effectively do in-context reasoning. Contrastive learning primarily focuses on discriminating positive pairs from negative pairs based on a \emph{global} semantic perspective. Auto-regressive models, on the other hand, accumulate their ability through \emph{local} token prediction. 
During inference, LLMs are expected to understand instruction, and jointly consider the logical relations between different hypothesises within single input.  By placing emphasis on fine-grained relations, the auto-regressive objective can better support in-context learning, enabling the model to grasp the nuanced connections and reasoning processes required for logical understanding.


Moreover, the auto-regressive objective significantly reduces computation costs during training by eliminating the need for negative candidates encoding. The streamlining of training process leads to more efficient and resource-friendly training without sacrificing performance. We also add another experiment by adjusting the ratio between counterfactual data and the normal ones as 1:1, and the comparison reveal that mixing more counterfactual data can also benefit the performance, which could be especially useful for low-resource domain, like finance and multi-lingual LLMs.

In summary, considering the advantages in both performance and training cost, the auto-regressive variant proves to be a superior choice for incorporating logic reasoning into LLMs.

\subsection{Factors Relevant to Logic Prior}

In Table~\ref{tab:ablation}, we also present the ablation results on \llama-33B when the counterfactual data augmentation strategy is omitted. Without the inclusion of counterfactual data, \meritp~degrades into a conditional generative task that can be solved through memorization, as each sample has its own prototypes within Wikipedia.

As indicated in the table, even without the augmentation (\textit{no aug.}), \meritp~still contributes to the enhancement of logical reasoning abilities, albeit with more limited improvements. However, the introduction of counterfactual data augmentation to eliminate memorization effects can further amplify the benefits. The overall experimental results point out that relation construction serves as effective supervision signal for introducing logic prior. 
We leave the work about developing novel techniques to prevent memorization but less involve factual noise as future work.

\begin{table}[t]
\centering
\setlength{\tabcolsep}{2.0mm}{
\scalebox{0.8}{
\begin{tabular}{lcccc}
\toprule
                        & \multicolumn{2}{c}{\textbf{ReClor}}  & \multicolumn{2}{c}{\textbf{LogiQA-v2}}\\
Model / Dataset         & Dev         & Test             & Dev          & Test               \\
\hline
\textit{\flant-3B} \\
~~~~w/ FLAN             & 53.6        & 53.8             & 49.5         & 49.5       \\
~~~~w/ \meritp~\& FLAN  & \bf{55.8}   & \bf{54.1}        & \bf{50.8}    & \bf{50.1}     \\ 
  \hdashline
\textit{\flant-11B} \\
~~~~w/ FLAN             & 58.0        & 60.5             & \bf{56.9}    & 53.6        \\
~~~~w/ \meritp~\& FLAN  & \bf{61.2}   & \bf{61.1}        & 56.0         & \bf{54.0}   \\ \hline
\textit{\llama-13B}       \\
~~~~w/ GPT4ALL          & 37.4        & 36.1             & 37.2         & 34.3          \\ 
~~~~w/ \meritp~\& GPT4All& \bf{39.2}  & \bf{37.7}        & 37.2         & \bf{35.1}        \\
\bottomrule
\end{tabular}
}}
\caption{Ablation study to explore if \meritp~can be combined with instruction tuning. For \flant~, we  use the subset of FLAN collection. For \llama, we introduce GPT4All~\citep{gpt4all}. 
}
\vspace{-0.3cm}
\label{tab:instruction-tuning}
\end{table}

\subsection{Compatibility with Instruction Tuning}

Instruction tuning has served as a critical step to make LLMs better in following human instruction, and/or generating with less toxic. In this section, we hope to study if \meritp~can be well integrated with supervised instruction tuning so that \meritp~has the potential to serve as a basic approach to train logic-enhanced foundation model before building applications. 
For \flant, we directly use the same subset of FLAN collection with our approach as the instruction tuning data. For \llama~models, we introduce GPT4All~\citep{gpt4all} data for extra supervision. During training, we simply sum the loss of instruction tuning and \meritp~in multitask training manner to keep the same data ratio.

As shown in Table~\ref{tab:instruction-tuning}, on most dataset splits, \meritp~can achieve additional improvements compared with the instruction tuning-only baselines. Specifically, we find that the improvements are more significant on ReClor that those on LogiQA-v2. One possible reason is that the language style in LogiQA-v2 is more close to formal language, leaving a gap with the natural user questions.

\subsection{Data Assumption Auto-Verification}

\begin{table}[t]
    \centering
    \setlength{\tabcolsep}{1.5mm}{
    \scalebox{0.72}{
    \begin{tabular}{c|cccc}
    \toprule
       Model     & Normal & Normal (Anony.) & C.F. & C.F. (Anony.)\\
       \midrule
       ChatGPT   & 94\%   & 77.4\% (-16.6\%)& 49.2\% & 65.0\% (+14.8\%) \\
       GPT-4     & 99.8\% & 99.2\% (-0.6\%) & 71.4\% & 94.2\% (+22.8\%) \\
    \bottomrule
    \end{tabular}
    }}
    \caption{The ratio of consistent data deemed by ChatGPT and GPT-4. \emph{Anony.} refers to \emph{anonymization} and \emph{C.F.} is the simplification of \emph{Counterfactual}.}
    \label{tab:verify}
\end{table}

In order to verify the rationality of our assumption that the \emph{direct} and \emph{indirect} relations are logically consistent, we employ ChatGPT and GPT-4 for automatic evaluations.
Specifically, we randomly sample 1,000 examples from the development set for our pre-training with the ratio of normal data and counterfactual ones as 1:1. For each data pair, we ask ChatGPT/GPT-4 to determine if the relation between the target entities are logically consistent. The prompt we used is shown in Appendix~\ref{sec:verify}. 
We have involved four different settings. Beside the normal data and the counterfactual ones, we have also applied anonymization~\citep{gcc2020kdd-qiu} to them to decouple the background knowledge from entity. Specifically, the target entities are replaced with [X] and [Y], and for counterfactual data, the other replaced entities during data augmentation are not further anonymized. Some cases can also be found in Appendix~\ref{sec:verify} for clearer understanding.


Our results are shown in Tabel~\ref{tab:verify}, from which we can observe that:
(1) for normal data, ChatGPT and GPT-4 deem that the logically consistent data occupie high ratios, which has initially verified the rationality of our data construction assumption. 
(2) For counterfactual data, the ratios significantly decrease. Yet, in the view of GPT-4, there is still more than 70\% of logically consistent data in the whole corpus. 
(3) When combined with entity anonymization, the ratios become much higher for counterfactual data, i.e., nearly 15\% absolute improvements for ChatGPT and 23\% for GPT-4. Besides, the ratio of normal data decreases significantly for ChatGPT, but is less perturbed for GPT-4.
The observation further demonstrates that most counterfactual data should also hold the assumption since the anonymization only remove the backgrounds of entities, yet leaving the context as original. And the great variation brought by counterfactual data augmentation also reveals the potential weakness of current LLMs on identifying the true causal relations.

\begin{figure}[!t]
    \centering
    \setlength{\abovecaptionskip}{0.12cm}
    \includegraphics[width=0.95\linewidth]{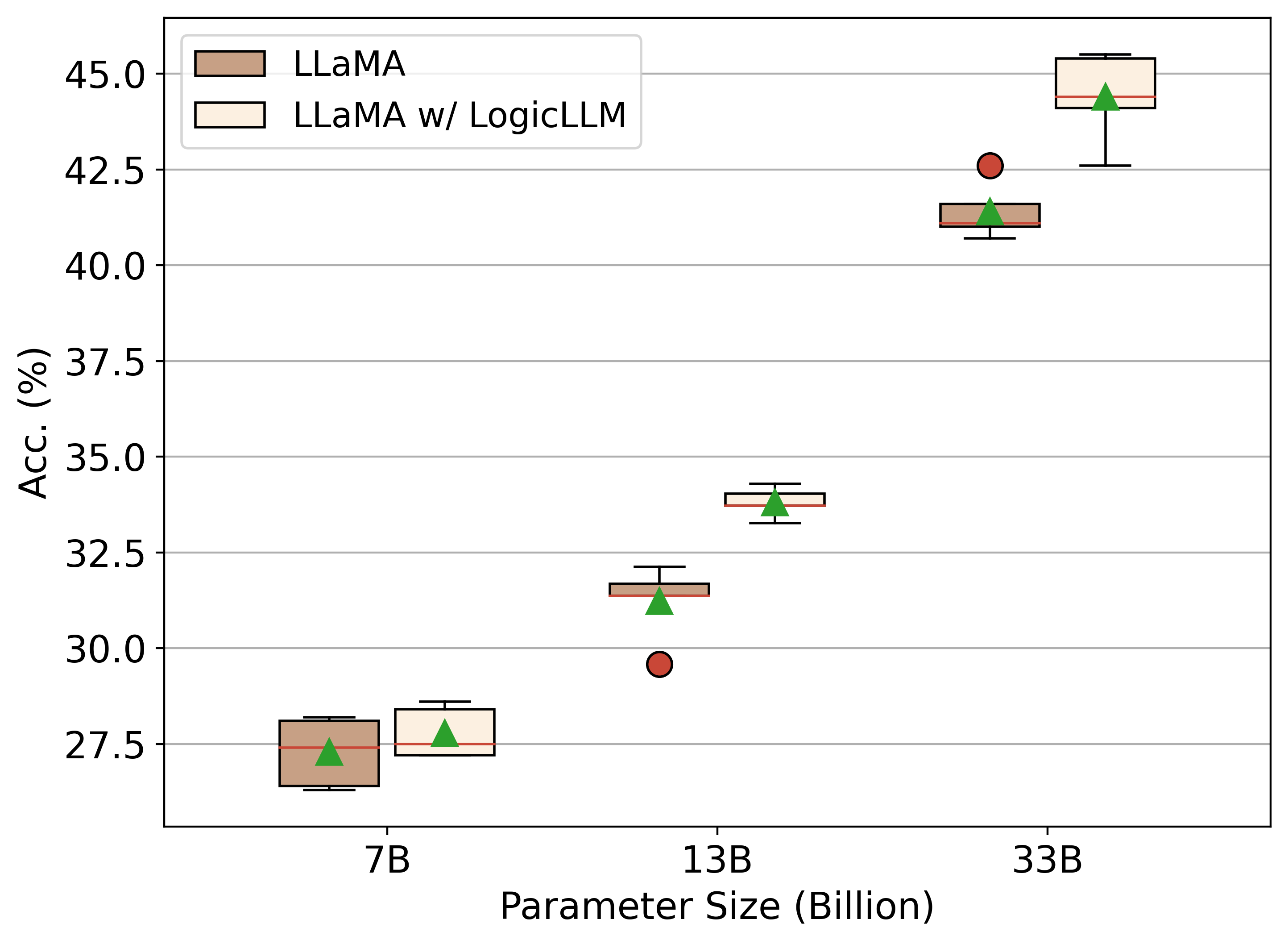}
    \caption{Results of 5 experiments with different option input orders across different model sizes on the test set of LogiQA-v2. \textbf{Brown circular marker}: outlier, \textbf{green triangle}: arithmetic mean value. 
    }
    \label{fig:robust}
\end{figure}
\subsection{Robustness}
By training LLMs on logic-consistent data and counterfactual augmentations, they are exposed to a wide range of input variations. This exposure helps them become less sensitive to minor perturbations such as shuffling of input options.  
To determine the robustness of \meritp~, we conducted experiments on LogiQA-v2 using models of varying sizes.  We shuffled the input order of different options and reperformed the inference process.

Figure~\ref{fig:robust} illustrates the findings of our experiments. We observed that \llama~exhibited higher variance across different input option orders, as indicated by the greater spread in results. The circular outlier values that indicate specific input orders causing significant variations, leading to substantially higher or lower performance results. Our observation is consistent with the recent findings of ~\citet{position-bias2023wang}, suggesting that the normal LLMs heavily suffer from position bias. 
In contrast, when \llama~is enhanced with \meritp, it achieves more stable performance across different parameter sizes. Moreover, the averaged performance of \llama~w/ \meritp~is significantly superior to that of \llama~alone. These results show that \meritp~ produces consistent and improved results compared to traditional LLMs, demonstrating the value of incorporating logic-enhanced training techniques into LLMs.

\begin{figure}
    \centering
    \setlength{\abovecaptionskip}{0.1cm}
    \includegraphics[width=\linewidth]{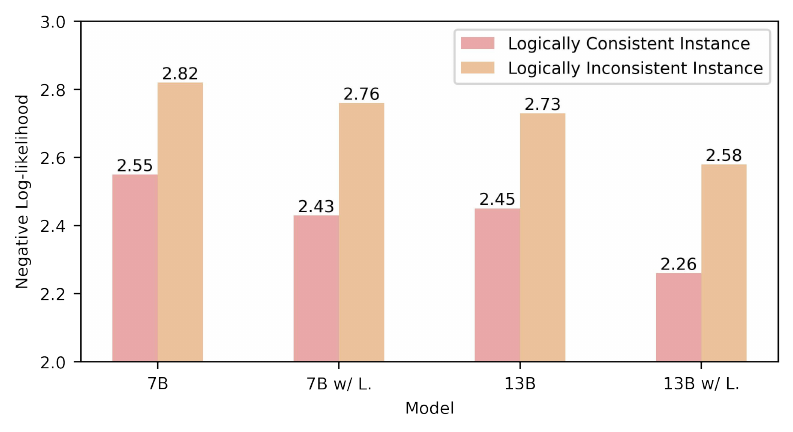}
    \caption{The averaged log-likelihood value of different models on the self-constructed logically consistent and inconsistent instances, respectively. w/ L. refers to the models augmented with \meritp.}
    \label{fig:lm-loss}
    \vspace{-0.1cm}
\end{figure}
\subsection{Training Quality Analysis}
\label{sec:quality}
In order to analyze the quality of our meta-training, we have constructed a test set using the framework of MERIt~\citep{merit}, which contains both logically consistent and inconsistent data. We have measured the log-likelihood on each sample as illustrated by Equation~\ref{eqn:logic}, and report the averaged results in Figure~\ref{fig:lm-loss}.

As shown in the figure, for logically consistent data, \meritp\ significantly reduced the negative log-likelihood. Moreover, the 7B-based model with \meritp\ surpasses the performance of \llama-13B. Notably, the disparity between the negative log-likelihood of logically consistent and inconsistent instances is further amplified, highlighting the effectiveness of \meritp\ in logical relation reconstruction.
Furthermore, our experiments suggest a decrease in the negative log-likelihood for logically inconsistent data. This observation exposes a weakness in the contrastive learning-based method, i.e., \merit, wherein the heuristic process for generating negative candidates introduces considerable noise. Consequently, some negative instances may not genuinely present contradictory logical relations.

\section{Conclusion}


In this paper, we have explored the feasibility and effectiveness of enhancing logical reasoning of LLMs via purely self-supervised training. We evaluate the performance based on two LLM series, i.e., \flant~and~\llama. The experimental results on two logical reasoning benchmarks, LogiQA-v2 and ReClor, demonstrate the effectiveness of our method. And the performance on RACE, MMLU and Big-Bench-Hard have also verified that the framework do not hurt the generalization of LLMs. Finally, we have analyzed the factors relevant to logic during training, and the compability with supervised instruction tuning. We hope the analysis could bring new insights to future research.

\section*{Acknowledgements}
This research is supported by the Ministry of Education, Singapore, under its Science of Learning Grant (award ID: MOE-MOESOL2021-0006). Any opinions, findings and conclusions or recommendations expressed in this material are those of the author(s) and do not reflect the views of the Ministry of Education, Singapore.

Besides, we sincerely appreciate the valuable comments from all the reviewers to help us make the paper polished.
We also greatly thank to Chengwei Qin and Professor Aixin Sun for their kind suggestions.

\section*{Limitations}

In this paper, we have explored the feasibility to introduce logical reasoning capability into LLMs via purely self-supervised meta-training. Though the results have demonstrated significant improvements on logical reasoning benchmarks, there are also some limitations:

\noindent\textbf{Randomness from Diverse Prompt/Instruction.} In our experiments, we find that the performance of LLMs, especially those never optimized by instruction tuning, is varying to different prompts. We try to reduce the variance by (1) using simpler prompt (as shown in Section~\ref{sec:prompt-template} or (2) using the released prompt by commonly accepted benchmark or leaderboard, e.g., MMLU, Big-Bench-Hard and Chain-of-Thought Hub~\citep{cot-hub}. Nevertheless, this still cannot entirely keep the certainty of the experimental results.

\noindent\textbf{Non-uniform Evaluation Strategy.} Currently, there is no \textit{de facto} technical standard for LLMs evaluation. Some work just let language models generate the response and match the content. However, this can be unfair for non-instruction-tuned models since they often cannot generate meaningful and complete sentences, especially those under 13 billion parameters. 

\noindent\textbf{Scaling.} Due to the resource limitation, we can only scale the method into models with 40 billion parameters under the help of low-rank adaptation.

\bibliography{anthology,custom}

\begin{thebibliography}{45}
\expandafter\ifx\csname natexlab\endcsname\relax\def\natexlab#1{#1}\fi

\bibitem[{Anand et~al.(2023)Anand, Nussbaum, Duderstadt, Schmidt, and Mulyar}]{gpt4all}
Yuvanesh Anand, Zach Nussbaum, Brandon Duderstadt, Benjamin Schmidt, and Andriy Mulyar. 2023.
\newblock \href {https://github.com/nomic-ai/gpt4all} {Gpt4all: Training an assistant-style chatbot with large scale data distillation from gpt-3.5-turbo}.

\bibitem[{Bang et~al.(2023)Bang, Cahyawijaya, Lee, Dai, Su, Wilie, Lovenia, Ji, Yu, Chung, Do, Xu, and Fung}]{bang2023chatgpt-eval}
Yejin Bang, Samuel Cahyawijaya, Nayeon Lee, Wenliang Dai, Dan Su, Bryan Wilie, Holy Lovenia, Ziwei Ji, Tiezheng Yu, Willy Chung, Quyet~V. Do, Yan Xu, and Pascale Fung. 2023.
\newblock \href {https://doi.org/10.48550/arXiv.2302.04023} {A multitask, multilingual, multimodal evaluation of chatgpt on reasoning, hallucination, and interactivity}.
\newblock \emph{CoRR}, abs/2302.04023.

\bibitem[{Chen et~al.(2022)Chen, Ma, Wang, and Cohen}]{pot2022chen}
Wenhu Chen, Xueguang Ma, Xinyi Wang, and William~W. Cohen. 2022.
\newblock \href {https://doi.org/10.48550/arXiv.2211.12588} {Program of thoughts prompting: Disentangling computation from reasoning for numerical reasoning tasks}.
\newblock \emph{CoRR}, abs/2211.12588.

\bibitem[{Cheng et~al.(2022)Cheng, Xie, Shi, Li, Nadkarni, Hu, Xiong, Radev, Ostendorf, Zettlemoyer, Smith, and Yu}]{binder2022cheng}
Zhoujun Cheng, Tianbao Xie, Peng Shi, Chengzu Li, Rahul Nadkarni, Yushi Hu, Caiming Xiong, Dragomir Radev, Mari Ostendorf, Luke Zettlemoyer, Noah~A. Smith, and Tao Yu. 2022.
\newblock \href {https://doi.org/10.48550/arXiv.2210.02875} {Binding language models in symbolic languages}.
\newblock \emph{CoRR}, abs/2210.02875.

\bibitem[{Chowdhery et~al.(2022)Chowdhery, Narang, Devlin, Bosma, Mishra, Roberts, Barham, Chung, Sutton, Gehrmann, Schuh, Shi, Tsvyashchenko, Maynez, Rao, Barnes, Tay, Shazeer, Prabhakaran, Reif, Du, Hutchinson, Pope, Bradbury, Austin, Isard, Gur{-}Ari, Yin, Duke, Levskaya, Ghemawat, Dev, Michalewski, Garcia, Misra, Robinson, Fedus, Zhou, Ippolito, Luan, Lim, Zoph, Spiridonov, Sepassi, Dohan, Agrawal, Omernick, Dai, Pillai, Pellat, Lewkowycz, Moreira, Child, Polozov, Lee, Zhou, Wang, Saeta, Diaz, Firat, Catasta, Wei, Meier{-}Hellstern, Eck, Dean, Petrov, and Fiedel}]{palm}
Aakanksha Chowdhery, Sharan Narang, Jacob Devlin, Maarten Bosma, Gaurav Mishra, Adam Roberts, Paul Barham, Hyung~Won Chung, Charles Sutton, Sebastian Gehrmann, Parker Schuh, Kensen Shi, Sasha Tsvyashchenko, Joshua Maynez, Abhishek Rao, Parker Barnes, Yi~Tay, Noam Shazeer, Vinodkumar Prabhakaran, Emily Reif, Nan Du, Ben Hutchinson, Reiner Pope, James Bradbury, Jacob Austin, Michael Isard, Guy Gur{-}Ari, Pengcheng Yin, Toju Duke, Anselm Levskaya, Sanjay Ghemawat, Sunipa Dev, Henryk Michalewski, Xavier Garcia, Vedant Misra, Kevin Robinson, Liam Fedus, Denny Zhou, Daphne Ippolito, David Luan, Hyeontaek Lim, Barret Zoph, Alexander Spiridonov, Ryan Sepassi, David Dohan, Shivani Agrawal, Mark Omernick, Andrew~M. Dai, Thanumalayan~Sankaranarayana Pillai, Marie Pellat, Aitor Lewkowycz, Erica Moreira, Rewon Child, Oleksandr Polozov, Katherine Lee, Zongwei Zhou, Xuezhi Wang, Brennan Saeta, Mark Diaz, Orhan Firat, Michele Catasta, Jason Wei, Kathy Meier{-}Hellstern, Douglas Eck, Jeff Dean, Slav Petrov, and Noah Fiedel.
  2022.
\newblock \href {https://doi.org/10.48550/arXiv.2204.02311} {Palm: Scaling language modeling with pathways}.
\newblock \emph{CoRR}, abs/2204.02311.

\bibitem[{Deng et~al.(2021)Deng, Su, Lees, Wu, Yu, and Sun}]{reason-bert}
Xiang Deng, Yu~Su, Alyssa Lees, You Wu, Cong Yu, and Huan Sun. 2021.
\newblock \href {https://doi.org/10.18653/v1/2021.emnlp-main.494} {Reasonbert: Pre-trained to reason with distant supervision}.
\newblock In \emph{{EMNLP}}, pages 6112--6127. {ACL}.

\bibitem[{Dettmers et~al.(2023)Dettmers, Pagnoni, Holtzman, and Zettlemoyer}]{dettmers2023qlora}
Tim Dettmers, Artidoro Pagnoni, Ari Holtzman, and Luke Zettlemoyer. 2023.
\newblock Qlora: Efficient finetuning of quantized llms.
\newblock \emph{CoRR}, abs/2305.14314.

\bibitem[{Fu et~al.(2023)Fu, Ou, Chen, Wan, Peng, and Khot}]{cot-hub}
Yao Fu, Litu Ou, Mingyu Chen, Yuhao Wan, Hao Peng, and Tushar Khot. 2023.
\newblock \href {https://doi.org/10.48550/arXiv.2305.17306} {Chain-of-thought hub: A continuous effort to measure large language models' reasoning performance}.
\newblock \emph{CoRR}, abs/2305.17306.

\bibitem[{Gao et~al.(2022)Gao, Madaan, Zhou, Alon, Liu, Yang, Callan, and Neubig}]{pal2022gao}
Luyu Gao, Aman Madaan, Shuyan Zhou, Uri Alon, Pengfei Liu, Yiming Yang, Jamie Callan, and Graham Neubig. 2022.
\newblock \href {https://doi.org/10.48550/arXiv.2211.10435} {{PAL:} program-aided language models}.
\newblock \emph{CoRR}, abs/2211.10435.

\bibitem[{Gao et~al.(2020)Gao, Wu, Li, Joty, Hoi, Xiong, King, and Lyu}]{discern-gao-2020}
Yifan Gao, Chien{-}Sheng Wu, Jingjing Li, Shafiq~R. Joty, Steven C.~H. Hoi, Caiming Xiong, Irwin King, and Michael~R. Lyu. 2020.
\newblock \href {https://doi.org/10.18653/v1/2020.emnlp-main.191} {Discern: Discourse-aware entailment reasoning network for conversational machine reading}.
\newblock In \emph{{EMNLP}}, pages 2439--2449. {ACL}.

\bibitem[{Hendrycks et~al.(2021)Hendrycks, Burns, Basart, Zou, Mazeika, Song, and Steinhardt}]{mmlu2021}
Dan Hendrycks, Collin Burns, Steven Basart, Andy Zou, Mantas Mazeika, Dawn Song, and Jacob Steinhardt. 2021.
\newblock Measuring massive multitask language understanding.
\newblock In \emph{{ICLR}}. OpenReview.

\bibitem[{Huang et~al.(2021)Huang, Fang, Cao, Wang, and Liang}]{dagn-huang-2021}
Yinya Huang, Meng Fang, Yu~Cao, Liwei Wang, and Xiaodan Liang. 2021.
\newblock \href {https://doi.org/10.18653/v1/2021.naacl-main.467} {{DAGN:} discourse-aware graph network for logical reasoning}.
\newblock In \emph{{NAACL-HLT}}, pages 5848--5855. {ACL}.

\bibitem[{Jiao et~al.(2021)Jiao, Guo, Niu, Ji, Li, and Nie}]{rept}
Fangkai Jiao, Yangyang Guo, Yilin Niu, Feng Ji, Feng{-}Lin Li, and Liqiang Nie. 2021.
\newblock \href {https://doi.org/10.18653/v1/2021.findings-acl.13} {{REPT:} bridging language models and machine reading comprehension via retrieval-based pre-training}.
\newblock In \emph{Findings of {ACL/IJCNLP}}, pages 150--163. {ACL}.

\bibitem[{Jiao et~al.(2022)Jiao, Guo, Song, and Nie}]{merit}
Fangkai Jiao, Yangyang Guo, Xuemeng Song, and Liqiang Nie. 2022.
\newblock \href {https://doi.org/10.18653/v1/2022.findings-acl.276} {Merit: Meta-path guided contrastive learning for logical reasoning}.
\newblock In \emph{Findings of {ACL}}, pages 3496--3509. {ACL}.

\bibitem[{Jiao et~al.(2023)Jiao, Wang, Huang, Wang, and Tu}]{mt-chatgpt2023}
Wenxiang Jiao, Wenxuan Wang, Jen{-}tse Huang, Xing Wang, and Zhaopeng Tu. 2023.
\newblock \href {https://doi.org/10.48550/arXiv.2301.08745} {Is chatgpt {A} good translator? {A} preliminary study}.
\newblock \emph{CoRR}, abs/2301.08745.

\bibitem[{Jurafsky and Martin(2009)}]{speech-and-lang-process}
Daniel Jurafsky and James~H. Martin. 2009.
\newblock \href {http://aleph.bib.uni-mannheim.de/F/?func=find-b&request=285413791&find_code=020&adjacent=N&local_base=MAN01PUBLIC&x=0&y=0} {\emph{Speech and language processing}}, 2. ed., [pearson international edition] edition.
\newblock Prentice Hall series in artificial intelligence. Prentice Hall, Pearson Education International.

\bibitem[{Lai et~al.(2017)Lai, Xie, Liu, Yang, and Hovy}]{race}
Guokun Lai, Qizhe Xie, Hanxiao Liu, Yiming Yang, and Eduard~H. Hovy. 2017.
\newblock \href {https://doi.org/10.18653/v1/d17-1082} {{RACE:} large-scale reading comprehension dataset from examinations}.
\newblock In \emph{{EMNLP}}, pages 785--794. {ACL}.

\bibitem[{Laskar et~al.(2023)Laskar, Bari, Rahman, Bhuiyan, Joty, and Huang}]{maruf2023eval}
Md~Tahmid~Rahman Laskar, M~Saiful Bari, Mizanur Rahman, Md~Amran~Hossen Bhuiyan, Shafiq Joty, and Jimmy~Xiangji Huang. 2023.
\newblock A systematic study and comprehensive evaluation of chatgpt on benchmark datasets.
\newblock In \emph{{ACL}}. {ACL}.

\bibitem[{Liu et~al.(2022{\natexlab{a}})Liu, Liu, Cui, Duan, Zhou, and Zhang}]{logiqa2}
Hanmeng Liu, Jian Liu, Leyang Cui, Nan Duan, Ming Zhou, and Yue Zhang. 2022{\natexlab{a}}.
\newblock Logiqa2.0 dataset - logical reasoning in mrc and nli tasks.
\newblock \emph{TASLP}.

\bibitem[{Liu et~al.(2020)Liu, Cui, Liu, Huang, Wang, and Zhang}]{logiqa}
Jian Liu, Leyang Cui, Hanmeng Liu, Dandan Huang, Yile Wang, and Yue Zhang. 2020.
\newblock \href {https://doi.org/10.24963/ijcai.2020/501} {Logiqa: {A} challenge dataset for machine reading comprehension with logical reasoning}.
\newblock In \emph{{IJCAI}}, pages 3622--3628.

\bibitem[{Liu et~al.(2022{\natexlab{b}})Liu, Li, He, Bing, Joty, and Si}]{kmlm}
Linlin Liu, Xin Li, Ruidan He, Lidong Bing, Shafiq~R. Joty, and Luo Si. 2022{\natexlab{b}}.
\newblock \href {https://arxiv.org/abs/2111.10962} {Knowledge based multilingual language model}.
\newblock In \emph{{EMNLP}}, pages 1--13. {ACL}.

\bibitem[{Longpre et~al.(2023)Longpre, Hou, Vu, Webson, Chung, Tay, Zhou, Le, Zoph, Wei, and Roberts}]{flan-v2}
Shayne Longpre, Le~Hou, Tu~Vu, Albert Webson, Hyung~Won Chung, Yi~Tay, Denny Zhou, Quoc~V. Le, Barret Zoph, Jason Wei, and Adam Roberts. 2023.
\newblock \href {https://doi.org/10.48550/arXiv.2301.13688} {The flan collection: Designing data and methods for effective instruction tuning}.
\newblock \emph{CoRR}, abs/2301.13688.

\bibitem[{Mialon et~al.(2023)Mialon, Dess{\`{\i}}, Lomeli, Nalmpantis, Pasunuru, Raileanu, Rozi{\`{e}}re, Schick, Dwivedi{-}Yu, Celikyilmaz, Grave, LeCun, and Scialom}]{arm-survey-lecun}
Gr{\'{e}}goire Mialon, Roberto Dess{\`{\i}}, Maria Lomeli, Christoforos Nalmpantis, Ramakanth Pasunuru, Roberta Raileanu, Baptiste Rozi{\`{e}}re, Timo Schick, Jane Dwivedi{-}Yu, Asli Celikyilmaz, Edouard Grave, Yann LeCun, and Thomas Scialom. 2023.
\newblock \href {https://doi.org/10.48550/arXiv.2302.07842} {Augmented language models: a survey}.
\newblock \emph{CoRR}, abs/2302.07842.

\bibitem[{Mintz et~al.(2009)Mintz, Bills, Snow, and Jurafsky}]{mintz2009distant}
Mike Mintz, Steven Bills, Rion Snow, and Daniel Jurafsky. 2009.
\newblock \href {https://aclanthology.org/P09-1113} {Distant supervision for relation extraction without labeled data}.
\newblock In \emph{Proceedings of the Joint Conference of the 47th Annual Meeting of the {ACL} and the 4th International Joint Conference on Natural Language Processing of the {AFNLP}}, pages 1003--1011. {ACL}.

\bibitem[{OpenAI(2023)}]{openai2023gpt}
OpenAI. 2023.
\newblock Gpt-4 technical report.
\newblock \emph{Preprint}.

\bibitem[{Ouyang et~al.(2022)Ouyang, Wu, Jiang, Almeida, Wainwright, Mishkin, Zhang, Agarwal, Slama, Ray, Schulman, Hilton, Kelton, Miller, Simens, Askell, Welinder, Christiano, Leike, and Lowe}]{instructgpt}
Long Ouyang, Jeff Wu, Xu~Jiang, Diogo Almeida, Carroll~L. Wainwright, Pamela Mishkin, Chong Zhang, Sandhini Agarwal, Katarina Slama, Alex Ray, John Schulman, Jacob Hilton, Fraser Kelton, Luke Miller, Maddie Simens, Amanda Askell, Peter Welinder, Paul~F. Christiano, Jan Leike, and Ryan Lowe. 2022.
\newblock \href {https://doi.org/10.48550/arXiv.2203.02155} {Training language models to follow instructions with human feedback}.
\newblock \emph{CoRR}, abs/2203.02155.

\bibitem[{Penedo et~al.(2023)Penedo, Malartic, Hesslow, Cojocaru, Cappelli, Alobeidli, Pannier, Almazrouei, and Launay}]{Falcon}
Guilherme Penedo, Quentin Malartic, Daniel Hesslow, Ruxandra Cojocaru, Alessandro Cappelli, Hamza Alobeidli, Baptiste Pannier, Ebtesam Almazrouei, and Julien Launay. 2023.
\newblock \href {https://doi.org/10.48550/arXiv.2306.01116} {The refinedweb dataset for falcon {LLM:} outperforming curated corpora with web data, and web data only}.
\newblock \emph{CoRR}, abs/2306.01116.

\bibitem[{Qin et~al.(2023)Qin, Zhang, Zhang, Chen, Yasunaga, and Yang}]{qin2023chatgpt-eval}
Chengwei Qin, Aston Zhang, Zhuosheng Zhang, Jiaao Chen, Michihiro Yasunaga, and Diyi Yang. 2023.
\newblock \href {https://doi.org/10.48550/arXiv.2302.06476} {Is chatgpt a general-purpose natural language processing task solver?}
\newblock \emph{CoRR}, abs/2302.06476.

\bibitem[{Qin et~al.(2021)Qin, Lin, Takanobu, Liu, Li, Ji, Huang, Sun, and Zhou}]{qin-erica}
Yujia Qin, Yankai Lin, Ryuichi Takanobu, Zhiyuan Liu, Peng Li, Heng Ji, Minlie Huang, Maosong Sun, and Jie Zhou. 2021.
\newblock \href {https://doi.org/10.18653/v1/2021.acl-long.260} {{ERICA}: Improving entity and relation understanding for pre-trained language models via contrastive learning}.
\newblock In \emph{{ACL/IJCNLP}}, pages 3350--3363. {ACL}.

\bibitem[{Qiu et~al.(2020)Qiu, Chen, Dong, Zhang, Yang, Ding, Wang, and Tang}]{gcc2020kdd-qiu}
Jiezhong Qiu, Qibin Chen, Yuxiao Dong, Jing Zhang, Hongxia Yang, Ming Ding, Kuansan Wang, and Jie Tang. 2020.
\newblock \href {https://doi.org/10.1145/3394486.3403168} {{GCC:} graph contrastive coding for graph neural network pre-training}.
\newblock In \emph{{KDD}}, pages 1150--1160. {ACM}.

\bibitem[{Schick et~al.(2023)Schick, Dwivedi{-}Yu, Dess{\`{\i}}, Raileanu, Lomeli, Zettlemoyer, Cancedda, and Scialom}]{toolformer}
Timo Schick, Jane Dwivedi{-}Yu, Roberto Dess{\`{\i}}, Roberta Raileanu, Maria Lomeli, Luke Zettlemoyer, Nicola Cancedda, and Thomas Scialom. 2023.
\newblock \href {https://doi.org/10.48550/arXiv.2302.04761} {Toolformer: Language models can teach themselves to use tools}.
\newblock \emph{CoRR}, abs/2302.04761.

\bibitem[{Suzgun et~al.(2022)Suzgun, Scales, Sch{\"{a}}rli, Gehrmann, Tay, Chung, Chowdhery, Le, Chi, Zhou, and Wei}]{big-bench-hard}
Mirac Suzgun, Nathan Scales, Nathanael Sch{\"{a}}rli, Sebastian Gehrmann, Yi~Tay, Hyung~Won Chung, Aakanksha Chowdhery, Quoc~V. Le, Ed~H. Chi, Denny Zhou, and Jason Wei. 2022.
\newblock \href {https://doi.org/10.48550/arXiv.2210.09261} {Challenging big-bench tasks and whether chain-of-thought can solve them}.
\newblock \emph{CoRR}, abs/2210.09261.

\bibitem[{Touvron et~al.(2023)Touvron, Lavril, Izacard, Martinet, Lachaux, Lacroix, Rozi{\`e}re, Goyal, Hambro, Azhar et~al.}]{touvron2023llama}
Hugo Touvron, Thibaut Lavril, Gautier Izacard, Xavier Martinet, Marie-Anne Lachaux, Timoth{\'e}e Lacroix, Baptiste Rozi{\`e}re, Naman Goyal, Eric Hambro, Faisal Azhar, et~al. 2023.
\newblock Llama: Open and efficient foundation language models.
\newblock \emph{CoRR}, abs/2302.13971.

\bibitem[{Wang et~al.(2023{\natexlab{a}})Wang, Liu, Huang, Jiao, Ding, Aw, and Chen}]{seaeval2023}
Bin Wang, Zhengyuan Liu, Xin Huang, Fangkai Jiao, Yang Ding, Ai~Ti Aw, and Nancy~F. Chen. 2023{\natexlab{a}}.
\newblock \href {https://arxiv.org/abs/2309.04766} {Seaeval for multilingual foundation models: From cross-lingual alignment to cultural reasoning}.
\newblock \emph{CoRR}, abs/2309.04766.

\bibitem[{Wang et~al.(2023{\natexlab{b}})Wang, Li, Chen, Zhu, Lin, Cao, Liu, Liu, and Sui}]{position-bias2023wang}
Peiyi Wang, Lei Li, Liang Chen, Dawei Zhu, Binghuai Lin, Yunbo Cao, Qi~Liu, Tianyu Liu, and Zhifang Sui. 2023{\natexlab{b}}.
\newblock \href {https://doi.org/10.48550/arXiv.2305.17926} {Large language models are not fair evaluators}.
\newblock \emph{CoRR}, abs/2305.17926.

\bibitem[{Wang et~al.(2022)Wang, Zhong, Tang, Wei, Fan, Jiang, Zhou, and Duan}]{lreasoner}
Siyuan Wang, Wanjun Zhong, Duyu Tang, Zhongyu Wei, Zhihao Fan, Daxin Jiang, Ming Zhou, and Nan Duan. 2022.
\newblock \href {https://doi.org/10.18653/v1/2022.findings-acl.127} {Logic-driven context extension and data augmentation for logical reasoning of text}.
\newblock In \emph{{ACL}}, pages 1619--1629. {ACL}.

\bibitem[{Wang et~al.(2021)Wang, Gao, Zhu, Zhang, Liu, Li, and Tang}]{wikidata}
Xiaozhi Wang, Tianyu Gao, Zhaocheng Zhu, Zhengyan Zhang, Zhiyuan Liu, Juanzi Li, and Jian Tang. 2021.
\newblock \href {https://doi.org/10.1162/tacl\_a\_00360} {{KEPLER:} {A} unified model for knowledge embedding and pre-trained language representation}.
\newblock \emph{{TACL}}, 9:176--194.

\bibitem[{Wei et~al.(2022{\natexlab{a}})Wei, Bosma, Zhao, Guu, Yu, Lester, Du, Dai, and Le}]{flan}
Jason Wei, Maarten Bosma, Vincent~Y. Zhao, Kelvin Guu, Adams~Wei Yu, Brian Lester, Nan Du, Andrew~M. Dai, and Quoc~V. Le. 2022{\natexlab{a}}.
\newblock Finetuned language models are zero-shot learners.
\newblock In \emph{{ICLR}}. OpenReview.

\bibitem[{Wei et~al.(2022{\natexlab{b}})Wei, Tay, Bommasani, Raffel, Zoph, Borgeaud, Yogatama, Bosma, Zhou, Metzler, Chi, Hashimoto, Vinyals, Liang, Dean, and Fedus}]{emergent2022jason}
Jason Wei, Yi~Tay, Rishi Bommasani, Colin Raffel, Barret Zoph, Sebastian Borgeaud, Dani Yogatama, Maarten Bosma, Denny Zhou, Donald Metzler, Ed~H. Chi, Tatsunori Hashimoto, Oriol Vinyals, Percy Liang, Jeff Dean, and William Fedus. 2022{\natexlab{b}}.
\newblock \href {https://doi.org/10.48550/arXiv.2206.07682} {Emergent abilities of large language models}.
\newblock \emph{CoRR}, abs/2206.07682.

\bibitem[{Wong et~al.(2023)Wong, Grand, Lew, Goodman, Mansinghka, Andreas, and Tenenbaum}]{word-model2world-model}
Lionel Wong, Gabriel Grand, Alexander~K. Lew, Noah~D. Goodman, Vikash~K. Mansinghka, Jacob Andreas, and Joshua~B. Tenenbaum. 2023.
\newblock \href {https://doi.org/10.48550/arXiv.2306.12672} {From word models to world models: Translating from natural language to the probabilistic language of thought}.
\newblock \emph{CoRR}, abs/2306.12672.

\bibitem[{Xu et~al.(2023)Xu, Sun, Zheng, Geng, Zhao, Feng, Tao, and Jiang}]{wizardlm2023xu}
Can Xu, Qingfeng Sun, Kai Zheng, Xiubo Geng, Pu~Zhao, Jiazhan Feng, Chongyang Tao, and Daxin Jiang. 2023.
\newblock Wizardlm: Empowering large language models to follow complex instructions.
\newblock \emph{CoRR}, abs/2304.12244.

\bibitem[{Xu et~al.(2021)Xu, Chen, and Zhao}]{dis-rel-ext}
Wang Xu, Kehai Chen, and Tiejun Zhao. 2021.
\newblock \href {https://doi.org/10.18653/v1/2021.findings-acl.144} {Discriminative reasoning for document-level relation extraction}.
\newblock In \emph{Findings of {ACL}}, pages 1653--1663. {ACL}.

\bibitem[{Yu et~al.(2020)Yu, Jiang, Dong, and Feng}]{reclor}
Weihao Yu, Zihang Jiang, Yanfei Dong, and Jiashi Feng. 2020.
\newblock \href {https://openreview.net/forum?id=HJgJtT4tvB} {Reclor: {A} reading comprehension dataset requiring logical reasoning}.
\newblock In \emph{{ICLR}}. OpenReview.

\bibitem[{Zeng et~al.(2021)Zeng, Wu, and Chang}]{sire}
Shuang Zeng, Yuting Wu, and Baobao Chang. 2021.
\newblock {SIRE:} separate intra- and inter-sentential reasoning for document-level relation extraction.
\newblock In \emph{Findings of {ACL}}, pages 524--534. {ACL}.

\bibitem[{Zhong et~al.(2021)Zhong, Wang, Tang, Xu, Guo, Wang, Yin, Zhou, and Duan}]{ar-lsat}
Wanjun Zhong, Siyuan Wang, Duyu Tang, Zenan Xu, Daya Guo, Jiahai Wang, Jian Yin, Ming Zhou, and Nan Duan. 2021.
\newblock \href {https://arxiv.org/abs/2104.06598} {{AR-LSAT:} investigating analytical reasoning of text}.
\newblock \emph{CoRR}, abs/2104.06598.

\end{thebibliography}

\newpage
\appendix
\begin{table}[h]
\centering
\setlength{\tabcolsep}{2.7mm}{
\scalebox{0.8}{
\begin{tabular}{lcccc}
\toprule
                        & \multicolumn{2}{c}{\textbf{ReClor}}   & \multicolumn{2}{c}{\textbf{LogiQA-v2}} \\
Model / Dataset         & Dev         & Test       & Dev        & Test \\
                        & Acc.        & Acc.       & Acc.       & Acc.     \\ \hline
\rob-L.                 & 62.6        & 55.6       & 59.8       & 57.0       \\
MERIt (RoBERTa-L)       & 69.4        & 61.6       & \tb{62.6}  & \tb{59.3}   \\
MERIt (DeBERTa-XXL)     & \tb{80.6}   & \tb{78.1}  & ---        & ---    \\ \hline
LLaMA-7B                & 28.8        & 28.3       & 24.4       & 23.7       \\
LLaMA-13B               & 31.6        & 34.4       & 31.6       & 31.1    \\ 
\llama-33B              & 45.2        & 50.3       & 41.2       & 41.6          \\\hline
GPT-3.5-turbo           & 56.6        & 61.2       & 54.5       & 52.7      \\ 
~~w/ CoT                & 58.8        & 57.7       & ---        & 53.1      \\ 
\bottomrule
\end{tabular}
}}
\caption{The overall accuracy of LLMs, i.e., ChatGPT (GPT-3.5-turbo) and LLaMA, and existing state-of-the-art methods~\cite{merit} on logical reasoning benchmarks. The evaluation of LLMs follows zero-shot in-context learning setting, where the models are expected to decode the answer based on the given instruction, context, and question.}
\label{tab:pre-exp}
\vspace{-0.4cm}
\end{table}

\section{Implementation Details}
\label{sec:app-imple}
\subsection{LLM Prompting}
\label{sec:llm-prompting}

In order to evaluate the generalization capabilities of LLMs across different tasks after post-training, we adopt a prompting-based approach. Here, the input to the LLMs is structured as \texttt{Instruction [Exemplars] Task input}. The instruction is tailored to the specific task at hand, while exemplars are utilized only in a few-shot setting. Each exemplar comprises both the task input and its corresponding output. For tasks such as multiple-choice question answering, the task input is a concatenation of the context, the question, and all potential options. The correct option index is used as the output. Besides, in a Chain-of-Thought (CoT) setting, we include a reasoning process formulated in natural language between the task input and output.

\subsection{Data}

We have constructed our self-supervised logic-enhanced training data from Wikipedia, where we directly used the paragraph corpus pre-processed by \citet{qin-erica}. We have constructed around 200K logically consistent sample pairs. After that, we further performed counterfactual data augmentation with the ratio of 1:3, and finally induced 800K training sample pairs in total. The data construction process mainly follows the original setting of \citet{merit} except two differences. First, we remove the usage of knowledge graph for relation annotation to enable fully self-supervision and simplify the construction workflow. Secondly, we have dropped the negative candidates since we employed auto-regressive training.

For language modeling, we employed different dataset with respect to the data used in their last stage training. For \flant~series models, we used the subset of FLAN-collection-v2~\citep{flan-v2};
while for \llama~series models, we used the same Wikipedia paragraphs from the corpus of \citet{qin-erica}. 

\subsection{Hyper-parameters of Training}
\label{sec:hyper-param}

During the pre-training process, we set the batch size to 4,096, which is implemented using gradient accumulation. The maximum sequence length is truncated at 1,024 for the FLAN collection and 512 for the \merit~corpus.
For the \flant~series models, we conduct training steps for 200 iterations, while for the \llama~series models, we perform training steps for 500 iterations. The learning rates are set as follows: 1e-4 for \flant-3B, 5e-5 for \flant-11B, 1e-5 for \llama-7B, and 5e-6 for \llama-13B.
To carry out the training process, we utilize 8 NVIDIA A100 80G GPUs. However, due to hardware limitations, models larger than 13B are trained using QLoRA~\citep{dettmers2023qlora}, a low-rank adaptation approach specifically designed for quantized LLMs. We follow the setting used in QLoRA with $\alpha$ as 16 and $r$ as 64. All linear layers are used for adaptation and the LoRA dropout is 0.05. The learning rate for \llama-33B and Falcon-40B is set as 5e-4.

\subsection{Evaluation}
To ensure a fair comparison, we maintain consistency across different models for each dataset. This involves using identical instructions and few-shot samples.
We use accuracy as the evaluation metric across all experiments.
The prompts for different dataset can be found in Appendix~\ref{sec:prompt-template}.

\begin{table}[t]
\centering
\setlength{\tabcolsep}{2.5mm}{
\scalebox{0.9}{
\begin{tabular}{lcccc}
\toprule
                        & High          & Middle        & Weighted     \\ \hline
\llama-7B                  & 46.9          & 61.1          & 51.0          \\
\llama-7B (Ours)           & ---           & ---           & 32.3          \\
\hdashline
\llama-13B                 & 47.2           & 61.6         & 51.4           \\
\llama-13B (Ours)          & ---           & ---           & \bf{54.5}           \\
\hdashline
\llama-33B                  & 48.3          & 64.1          & 52.9      \\
\llama-33B (Ours)           & ---           & ---           & \bf{68.1}      \\ 
\bottomrule
\end{tabular}
}}
\caption{The comparison on RACE dataset between our reproduced results and those reported by the opriginal paper of \llama.}
\label{tab:race-cmp}
\end{table}

\section{Interpretation for Different Results on RACE}
\label{sec:eval-strategy}

In this section, we will discuss the different results on RACE between ours and those reported by the original paper of LLaMA. Specifically, \citet{touvron2023llama} do not report the weighted results, so we convert them by ourselves. The results are shown in Table~\ref{tab:race-cmp}.
From the table we can find that only \llama-7B cannot match the performance reported by the authors. On \llama-13B and \llama-33B, our reproduced accuracies are much higher than the reported ones, which can help address the concern of unfair comparison, and demonstrate the effectiveness of our proposed \meritp.

\begin{table}[t]
\centering
\setlength{\tabcolsep}{2.0mm}{
\scalebox{0.8}{
\begin{tabular}{lcccc}
\toprule
Model / Dataset     & Zero-shot   & Direct     & CoT      \\ \hline
\llama-7B           & 24.9        & 30.4        & \bf 27.0\\

~~~~w/ \meritp       & \bf 25.2        & \bf{30.8}        & 25.9 \\ 
\hdashline
\llama-13B          & 25.0        & 34.7        & 32.3        \\
~~~~w/ \meritp       & \bf{26.3}   & \bf{35.0}   & \bf{33.9}   \\ %
\hline 
\flant-3B          & 38.0        & 40.2       & 35.1               \\
~~~~w/ \meritp~\& FLAN    & \bf{40.5}   & \bf{41.2}  & \bf{36.7}   \\ 
\hdashline
\flant-11B         & 43.0        & \bf 42.6       & \bf 40.9    \\
~~~~w/ \meritp~\& FLAN    & \bf{44.1}   & 36.2       & 40.2     \\  
\bottomrule
\end{tabular}
}}
\caption{The accuracy of \llama~and \flant~based models on BIG-Bench-Hard. \textit{Direct} refer to few-shot setting through direct prompting, where only the final answer is given. Instead, in \textit{CoT} setting, the reasoning process is also concatenated. The exemplars used for direct few-shot prompting and CoT prompting are consistent in each task, which are officially provided. 
}
\label{tab:bbh}
\vspace{-0.3cm}
\end{table}


\section{Logic-enhanced Meta-training for Complex Task Understanding}

We evaluated the performance of logic-enhanced pre-trained models on BIG-Bench-Hard, a benchmark comprising challenging tasks where human performance surpasses that of LLMs. Table~\ref{tab:bbh} presents the results achieved by the \llama~ and \flant~models under three evaluation settings: zero-shot, direct few-shot, and CoT.

In the zero-shot setting, our logic-enhanced meta-training significantly improves all four investigated models. For instance, the zero-shot accuracies of \llama-13B and \flant-T5-11B are 25.0\% and 38.0\%, respectively. When combined with the \meritp~model, the accuracy scores of \llama-13B and \flant-11B improve to 26.3\% and 44.1\%, respectively. 
Some tasks included in BBH require free-form answers thus we cannot evaluate the models by selecting the candidate with lowest perplexity or log likelihood. Instead, we need to follow the evaluation of API-based models, which employs regularization expression to capture the answer from the response.
However, smaller language models, especially those without being instruction tuned, fail to accept diverse instruction, and generate structured response.
As a result, the absolute performance under zero-setting setting of \llama-based models are relatively limited.

On the other hand, the direct few-shot results outperform the zero-shot results in three out of four models, with the exception of \flant-11B. Similarly, logic-enhanced meta-training boosts the performance of models, except for \flant-11B. In the CoT setting, our method further enhances the performances of \llama-13B and \flant-3B. However, the best direct few-shot and CoT results (42.6\% and 40.9\%, respectively) are both inferior to the best zero-shot result (44.1\%). Notably, the CoT results on \flant-3B are significantly worse than the zero-shot and direct few-shot results. 
These observations suggest the potential drawback that learning CoT from annotated training data, i.e., FLAN collection, has difficulty in generalizing to different task categories, for example, learning CoT from math word problem solving and solving logical puzzles.
We provide further discussion on these findings in Appendix~\ref{sec:weakness}.

\section{Prompt Template}
\label{sec:prompt-template}

\subsection{ReClor}

\begin{quote}

\texttt{\small{Answer the following question with the given context through logical reasoning:}}

\texttt{\small{Context: \#\textit{Context}}}

\texttt{\small{Question: \#\textit{Question}}}

\texttt{\small Options:\\
A: \#\textit{Option A}. \\
B: \#\textit{Option B}. \\
C: \#\textit{Option C}. \\ 
D: \#\textit{Option D}. }

\texttt{\small The answer is }

\end{quote}

\subsection{LogiQA-v2 \& RACE}

\begin{quote}

\texttt{\small{Answer the following question with the given context:}}

\texttt{\small Context: \#\textit{Context}}

\texttt{\small Question: \#\textit{Question}}

\texttt{\small Options:\\
A: \#\textit{Option A}. \\
B: \#\textit{Option B}. \\
C: \#\textit{Option C}. \\ 
D: \#\textit{Option D}. }

\texttt{\small The answer is }

\end{quote}

\subsection{MMLU}

\begin{quote}
\texttt{\small{The following are multiple choice questions (with answers) about \#\textit{Subject}.}}

\texttt{\small\#\textit{Question}}

\texttt{\small A: \#\textit{Option A}. \\
B: \#\textit{Option B}. \\
C: \#\textit{Option C}. \\ 
D: \#\textit{Option D}. }

\texttt{\small Answer: }

\end{quote}

\section{Auto-Verification Cases for Logical Consistency}
\label{sec:verify}

\subsection{Prompt Template}

\begin{quote}

\texttt{\small{[User]:}}  

\texttt{\small{Determine whether the relation between "[Entity A]" and "[Entity B]" in the given two sentences are logically consistent.}}  

\texttt{\small{Directly give the answer from either Yes or No.}}

\texttt{\small{Sentence 1:}}

\texttt{\small{[Sentence(s) 1]}}  
  
\texttt{\small{Sentence 2:}}  

\texttt{\small{[Sentence(s) 2]}}  
   
\texttt{\small{[ChatGPT/GPT-4]:}}

\texttt{\small{Yes/No.}}

\end{quote}

\subsection{Normal Version}

\begin{quote}

\texttt{\small{[User]:}}  

\texttt{\small{Determine whether the relation between "\textbf{Everdingen}" and "\textbf{Sweden}" in the given two sentences are logically consistent.}}  

\texttt{\small{Sentence 1:}}

\texttt{\small{In the manner of Frans Post, \textbf{Everdingen} took advantage of this mishap by making sketches of the Norwegian landscape, which would have seemed very exotic to his Dutch countrymen. His annotated drawings document visits to the south - east Norwegian coast and to Bohusland and the Göteborg area in western \textbf{Sweden}.}}  
  
\texttt{\small{Sentence 2:}}  

\texttt{\small{In 1644 \textbf{Everdingen} travelled to Norway and \textbf{Sweden}, a trip that was to have profound consequences on his art.}}  

\texttt{\small{The output should either be Yes or No.}}

\texttt{\small{[ChatGPT]:}}

\texttt{\small{Yes.}}

\end{quote}

\subsection{Counterfactual Version}

\begin{quote}

\texttt{\small{[User]:}}  

\texttt{\small{Determine whether the relation between "\textbf{Nicholas Roerich}" and "\textbf{Master}" in the given two sentences are logically consistent.}}  

\texttt{\small{Sentence 1:}}

\texttt{\small{In the manner of Frans Post, \textbf{Nicholas Roerich} took advantage of this mishap by making sketches of the Canal del Dique landscape, which would have seemed very exotic to his Dutch countrymen. His annotated drawings document visits to the south - east Canal del Dique coast and to Bohusland and the Göteborg area in western \textbf{Master}.}}  
  
\texttt{\small{Sentence 2:}}  

\texttt{\small{In 1644 Nicholas Roerich travelled to Norway and Master , a trip that was to have profound consequences on his art .}}  

\texttt{\small{The output should either be Yes or No.}}

\texttt{\small{[ChatGPT]:}}

\texttt{\small{No.}}

\end{quote}

Entity replacement:

\begin{itemize}
    \item Everdingen $\;\rightarrow\;$ Nicholas Roerich;
    \item Sweden $\;\rightarrow\;$ Master;
    \item Norwegian (connecting entity) $\;\rightarrow\;$ Canal del Dique;
\end{itemize}

\subsection{Anonymized Version}

\begin{quote}

\texttt{\small{[User]:}}  

\texttt{\small{Determine whether the relation between "\textbf{[X]}" and "\textbf{[Y]}" in the given two sentences are logically consistent.}}  

\texttt{\small{Sentence 1:}}

\texttt{\small{In the manner of Frans Post, \textbf{[X]} took advantage of this mishap by making sketches of the Canal del Dique landscape , which would have seemed very exotic to his Dutch countrymen. His annotated drawings document visits to the south - east Canal del Dique coast and to Bohusland and the Göteborg area in western \textbf{[Y]}.}}  
  
\texttt{\small{Sentence 2:}}  

\texttt{\small{In 1644 \textbf{[X]} travelled to Norway and \textbf{[Y]}, a trip that was to have profound consequences on his art .}}  

\texttt{\small{The output should either be Yes or No.}}

\texttt{\small{[ChatGPT]:}}

\texttt{\small{Yes.}}

\end{quote}

\begin{table}[t]
\centering
\setlength{\tabcolsep}{2.5mm}{
\scalebox{0.9}{
\begin{tabular}{lcccc}
\toprule
                        & \multicolumn{2}{c}{\textbf{ReClor}}  & \multicolumn{2}{c}{\textbf{LogiQA-v2}} \\
Model / Dataset         & Dev         & Test          & Dev     & Test \\
                        & Acc.        & Acc.          & Acc.    & Acc.     \\ \hline
\textit{zero-shot} \\ \hline
ChatGPT                 & 56.6        & 61.2          & 54.5    & 52.7      \\ 
~~w/ CoT                &  58.8        & 57.7          & 54.5    & 53.1      \\ \hline
\textit{5-shot}  \\ \hline
ChatGPT           & 61.0        & \textbf{63.0} & 55.1   & 54.5      \\
~~w/ CoT              & \bf{62.0}   & 62.5 & 47.6   & \bf{55.6}   \\
~~w/ CoT + Cate.      &  N/A        & N/A           & \bf{55.8}& 55.0           \\
\bottomrule
\end{tabular}
}}
\caption{The results on logical reasoning benchmarks with enhanced Chain-of-Thought prompting.}
\vspace{-0.3cm}
\label{tab:cot}
\end{table}

\section{Discussion about Different Perspectives of Logical Reasoning}

In our opinion, logic can be reflected through multiple aspects. Here, we use a simple logic rule to discuss the different perspectives:
\begin{equation}
    (\alpha\to\beta) \wedge (\beta\to\gamma) \leftrightarrow \alpha\to\gamma.
\end{equation}

The above equation shows the simplest case of first-order logic reasoning, where $\alpha,\,\beta$ and $\gamma$ are different variables, and $\wedge$ is \textit{logical and}. We can also introduce the necessary logical connectives in natural language to make it easier for understanding:
\begin{equation}
    \mathrm{IF}\;\alpha\to\beta\;\mathrm{AND}\;\beta\to\gamma,\;\mathrm{THEN}\;\alpha\to\gamma.
\end{equation}
It should be noted that, in symbolic logic, we often ignore the actual meaning of relations. However, we can always find a path, i.e., a series of relation triplets from knowledge graph to transform the above symbolic form into natural language based logical reasoning process:
\begin{equation}
    \mathrm{IF}\;\alpha\relto{r_1}\beta\;\mathrm{AND}\;\beta\relto{r_2}\gamma,\;\mathrm{THEN}\;\alpha\relto{r_3}\gamma.
\end{equation}
One example here can be: $r_1$ refers to \textit{is the father of}, $r_2$ refers to \textit{is the mother of}, and $r_3$ refers to \textit{is the grandpa of}.

From the above discussion, we can conclude that (1) logical connectives focus on discourse-level connections, (2) symbolic logic can be viewed as the simplified version of logical reasoning in natural language, where we focus more on the formal rules of atomic logic operations, and (3) relational reasoning concentrates on the actual logic operations built on world knowledge. Both of what we have discussed in the paper and the reviewers have mentioned in comments, i.e., logical connectives, are indeed different perspectives of logical reasoning. They do not contradict to each other, and discussing them separately is beneficial to make the problem easier. Besides, there are also several studies also discuss logical reasoning from the relational reasoning perspective~\citep{word-model2world-model,dis-rel-ext,sire,lreasoner}. And Figure~\ref{fig:logiqa-sample} also shows the case emphasizing relational reasoning.

\section{Weakness of LLMs on Logical Reasoning}
\label{sec:weakness}
Table~\ref{tab:cot} showcases the evaluation results of LLMs' performance in both few-shot and CoT settings. The intermediate reasoning process is automatically generated by ChatGPT using the prompt \textit{``Let's think step by step.''} In the case of zero-shot CoT, we include the suffix prompt \textit{``So the answer is''} to guide the models in summarizing and concluding the answer. For few-shot CoT, the reasoning process is initially generated for each sample in the training set. Subsequently, we retain the samples where the final prediction is correct, following the steps outlined in zero-shot CoT. During testing, we randomly select samples from the retained candidates, as well as the automatically generated CoT, to serve as exemplars. 

However, our observations indicate that both few-shot learning and the use of CoT do not significantly improve the models' performance. 
For example, ChatGPT w/ CoT performs much worse than that without CoT on the development set of LogiQA-v2.
One potential reason for this is that the selected samples differ substantially from the target example. To investigate further, we incorporate reasoning category information during exemplar selection. In LogiQA-V2, each question is annotated with a reasoning category, such as categorical reasoning, sufficient conditional reasoning, or necessary conditional reasoning. For few-shot CoT prompting, we only consider candidates that share at least two common reasoning categories. 
This particular variant is denoted as \textit{``ChatGPT w/ CoT + Cate.''} in the table.

Despite these efforts, we find that carefully selecting prompting exemplars only provides limited improvement. The results indicate that LLMs struggle to comprehend the reasoning structure from a limited number of observed examples. Consequently, they face challenges in effectively learning the mapping between input-label and input-rationale-label. Additionally, as shown in Table~\ref{tab:logic}, we observe that \meritp~also contributes minimally to addressing this issue. We recognize the need for further investigation in this area and leave it as a potential avenue for future research.




\end{document}